# Asymptotic Efficiency of Deterministic Estimators for Discrete Energy-Based Models: Ratio Matching and Pseudolikelihood

**Benjamin M. Marlin**
Department of Computer Science
University of British Columbia
Vancouver, Canada

**Nando de Freitas**
Department of Computer Science
University of British Columbia
Vancouver, Canada

## Abstract

Standard maximum likelihood estimation cannot be applied to discrete energy-based models in the general case because the computation of exact model probabilities is intractable. Recent research has seen the proposal of several new estimators designed specifically to overcome this intractability, but virtually nothing is known about their theoretical properties. In this paper, we present a generalized estimator that unifies many of the classical and recently proposed estimators. We use results from the standard asymptotic theory for M-estimators to derive a generic expression for the asymptotic covariance matrix of our generalized estimator. We apply these results to study the relative statistical efficiency of classical pseudolikelihood and the recently-proposed ratio matching estimator.

## 1 Introduction

The class of discrete energy-based models contains many important models in machine learning and statistics including Ising models [Ising, 1925], discrete Markov random fields [Kindermann et al., 1980], Boltzmann machines [Ackley et al., 1985], restricted Boltzmann machines [Smolensky, 1986], discrete exponential family harmoniums [Welling et al., 2005] and conditional random fields [Lafferty et al., 2001]. These models have a diverse range of applications in multiple fields including image modeling, computer vision, natural language processing, information filtering and retrieval, remote sensor networks and deep learning.

Unlike Bayesian networks [Pearl, 1988], these models are defined through arbitrary positive potential functions and must be explicitly normalized. The normalization step involves computing a sum over all possible joint states of the random variables in the model. The computation of the normalization term, or partition function, is tractable for certain limited graphical structures like chains or trees. In the general case, it requires an amount of computation that is exponential in the number of random variables in the model. As a result, the accurate and computationally efficient estimation of discrete energy-based models is a fundamental open problem in machine learning and computational statistics.

Maximum likelihood is the most commonly used estimator for probabilistic models. Its prevalence is based on two key properties: asymptotic consistency and maximal asymptotic efficiency [Fisher, 1922, p. 316].[1] Essentially, an estimator is asymptotically consistent if it converges to the true parameters as the sample size goes to infinity. An asymptotically consistent estimator is maximally efficient if the variance in the estimated parameters attains the minimum possible value among all consistent estimators as the sample size goes to infinity.[2]

However, the criterion function underlying standard maximum likelihood estimation requires the computation of normalized joint probabilities under the model, which in turn requires the explicit computation of the partition function. When the computation of the partition function is intractable, the computation of the maximum likelihood estimator is clearly also intractable. This creates a fundamental tension between classical estimation theory, which suggests the use of maximum likelihood estimation due to its optimal statistical properties, and complexity theory, which states that the computation required to perform maximum likelihood estimation is infeasible for a general discrete energy-based model.

---

[1] Note that these properties hold subject to particular regularity conditions that are satisfied by many model classes including discrete log-linear models like Markov random fields, Boltzmann machines and conditional random fields.

[2] Note that both properties implicitly assume that the model is well-specified (the true distribution can be represented by the model class), although the statements can be generalized when this is not the case.

The estimation of discrete energy-based models thus requires a trade-off between the statistical and computational properties of estimators. Two fundamentally different approaches have been proposed for this problem. The first approach is to approximately maximize the likelihood. Examples of the approximation approach include stochastic approximation-based maximum likelihood methods [Younes, 1989], as well as approximation methods based on belief propagation [Wainwright et al., 2003].

The second approach is to select an alternative estimation criterion that explicitly avoids the computation of the partition function. Pseudolikelihood is perhaps the best-known alternative estimation criterion for discrete energy-based models [Besag, 1975]. It avoids the need to compute the partition function by defining a criterion function consisting of a sum of log-conditional probabilities. Since any conditional probability can be written as a ratio of model probabilities with the same value of the partition function, the partition function analytically cancels away.

Recent research in machine learning has seen the proposal of a large number of alternative deterministic estimators for discrete energy-based models that resemble the original pseudolikelihood estimator proposed by Besag [1975], as well as various forms of composite likelihood estimators [Lindsay, 1988]. This set of alternative estimators includes ratio matching [Hyvärinen, 2007], generalized score matching [Lyu, 2009], contrastive estimation [Smith and Eisner, 2005] and non-local contrastive objectives [Vickrey et al., 2010]. While previous work has focused on proposing new estimators and empirical comparisons among existing estimators [Marlin et al., 2009], virtually nothing is known about the asymptotic relative efficiencies of these alternative estimators.

Our primary interest in this paper is the unification and theoretical analysis of deterministic estimators for discrete energy-based models. We present a novel generalized estimator and show that it subsumes the majority of the deterministic estimators we have mentioned thus far. Our generalized estimator is itself a special case of an M-estimator [van der Vaart, 2000]. This allows us to apply the standard asymptotic theory for M-estimators to obtain a generic expression for the asymptotic covariance matrix of our generalized estimator. We use these results to study the relative efficiency of pseudolikelihood and ratio matching. We present a theorem bounding the asymptotic covariance matrix for ratio matching by the asymptotic covariance matrix for pseudolikelihood. We show that neither of these estimators is strictly more efficient than the other.

## 2 Discrete Energy-Based Models

Discrete energy-based models fall into two broad categories: discriminative and generative models. We describe the discriminative case as it subsumes the generative case. A discriminative discrete energy-based model is a joint distribution over a set of random variables $\boldsymbol{X}$, conditioned on an instantiation of a second set of random variables $\boldsymbol{Y}$.[3] We assume that there are $D_X$ random variables $\boldsymbol{X} = (X_1, ..., X_{D_X})$ in the first set and $D_Y$ random variables $\boldsymbol{Y} = (Y_1, ..., Y_{D_Y})$ in the second set. We assume that each random variable $X_d$ takes values in a discrete set $\mathcal{X}_d$. The set of random variables $\boldsymbol{X}$ thus takes values in the set $\mathcal{X} = \mathcal{X}_1 \times ... \times \mathcal{X}_{D_X}$. The set of random variables $\boldsymbol{Y}$ similarly takes values in the set $\mathcal{Y} = \mathcal{Y}_1 \times ... \times \mathcal{Y}_{D_Y}$, but the domain of each variable $\mathcal{Y}_d$ is unconstrained.

The model is defined through an energy function $E_\theta(\boldsymbol{x}, \boldsymbol{y})$, which depends on the model parameters $\theta$ and a pair of instantiations $\boldsymbol{x} = (x_1, ..., x_{D_X})$, $\boldsymbol{y} = (y_1, ..., y_{D_Y})$. We denote the dimensionality of the parameter vector by $D_\theta$. The joint probability and partition function for the discriminative dEBM are shown in equations 2.1 and 2.2.

$$P_\theta(\boldsymbol{X} = \boldsymbol{x}|\boldsymbol{Y} = \boldsymbol{y}) = \frac{1}{\mathcal{Z}_\theta(\boldsymbol{y})} \exp\left(-E_\theta(\boldsymbol{x}, \boldsymbol{y})\right) \quad (2.1)$$

$$\mathcal{Z}_\theta(\boldsymbol{y}) = \sum_{\boldsymbol{x} \in \mathcal{X}} \exp\left(-E_\theta(\boldsymbol{x}, \boldsymbol{y})\right) \quad (2.2)$$

The generative dEBM can be seen as a special case of the discriminative dEBM where the set of conditioning variables $\boldsymbol{Y}$ is empty. In this case, we use the simplified notation $E_\theta(\boldsymbol{x})$ for the energy function and $\mathcal{Z}_\theta$ for the partition function. Unlike Bayesian networks [Pearl, 1988], energy-based models are not automatically normalized. The exponentiated energies must be explicitly divided by the partition function to yield a valid joint distribution. The partition function is a sum over all joint instantiations of the $\boldsymbol{X}$ random variables with the conditioning instantiation $\boldsymbol{y}$ fixed (or empty for the generative case). The number of terms in this sum grows exponentially as a function of $D_X$ and it quickly becomes intractable to compute for both discriminative and generative dEBMs. We illustrate several common instances of the dEBM framework below, including binary Markov random Fields (MRFs),

---

[3] We reverse the standard roles of $X$ and $Y$ in conditional models so that the conditional model simplifies to a generative model over the variables $X$ when the conditioning set it empty.

binary restricted Boltzmann machines (RBMs) and binary conditional random fields (CRFs).

- **Binary MRF:** The binary MRF model is parameterized by a symmetric $D_X \times D_X$ quadratic interaction matrix $W$ [Kindermann et al., 1980]. It is the discrete analogue of the multivariate Gaussian distribution.

$$E_\theta^{MRF}(\boldsymbol{x}) = -\boldsymbol{x}^T W \boldsymbol{x} \qquad (2.3)$$

- **Binary CRF:** A binary CRF model is parameterized by a symmetric $D_X \times D_X$ quadratic interaction matrix $W$ and a bi-linear interaction term linking the features $\boldsymbol{Y}$ to the labels $\boldsymbol{X}$. It is the discriminative analogue of the binary MRF model [Lafferty et al., 2001].

$$E_\theta^{CRF}(\boldsymbol{x}, \boldsymbol{y}) = -(\boldsymbol{x}^T W \boldsymbol{x} + \boldsymbol{x}^T V \boldsymbol{y}) \qquad (2.4)$$

- **Binary RBM:** The binary RBM model is parameterized by $K$ filter vectors $W_k$ of size $D_X \times 1$ and $K$ scalar offsets $c_k$. This form of the energy function is obtained by starting from a bipartite binary random field model, the more common visible/hidden representation of the RBM, and analytically summing out over the hidden variables [Smolensky, 1986].

$$E_\theta^{RBM}(\boldsymbol{x}) = -\sum_{k=1}^{K} \log(1 + \exp(\boldsymbol{x}^T W_k + c_k)) \qquad (2.5)$$

Perhaps the most significant difference between MRFs, CRFs and RBMs is that MRFs and CRFs have energy functions that are linear in their parameters (they are log-linear models), while the RBM does not. In addition, the RBM model is not identifiable due to the fact that permuting the $K$ filters leaves the model probability invariant. The estimators that we study in this paper are applicable to all dEBMs, but log-linear models are more amenable to theoretical analysis due to the fact that their log likelihoods are convex.

## 3 A Generalized Estimator

In this section, we introduce a novel generalized estimator that subsumes many of the classical and recently-proposed estimators for discrete energy-based models. In the first part of this section, we define the generalized estimator. The generalized estimator is itself a member of a larger family of estimators called *M-estimators* [van der Vaart, 2000, Ch. 5]. In the subsequent parts of this section, we derive a generic expression for the asymptotic covariance matrix of the generalized estimator using the standard asymptotic theory of M-estimators.

### 3.1 Definition of the Estimator

Our generalized estimator deals with the intractability of the partition function in exactly the same way as Besag's classical pseudolikelihood estimator [Besag, 1975]. It is based on the use of ratios of model probabilities with the same partition function, ensuring that the partition functions analytically cancel away. To emphasize the point that the estimator does not rely on the availability of normalized probabilities, we introduce the special notation $Q_\theta(\boldsymbol{x})$ in Equation 3.8 to denote the model probability computed up to the normalization term.

$$Q_\theta(\boldsymbol{x}) = \exp(-E_\theta(\boldsymbol{x})) \qquad (3.8)$$

$$R_\theta(\boldsymbol{x}, \mathcal{A}) = \frac{P_\theta(\boldsymbol{x})}{\sum_{\boldsymbol{x}' \in \mathcal{A}} P_\theta(\boldsymbol{x}')} = \frac{Q_\theta(\boldsymbol{x})}{\sum_{\boldsymbol{x}' \in \mathcal{A}} Q_\theta(\boldsymbol{x}')} \qquad (3.9)$$

We define the probability ratio function $R_\theta(\boldsymbol{x}, \mathcal{A})$ in Equation 3.9. This function takes an arbitrary data configuration $\boldsymbol{x} \in \mathcal{X}$ and an arbitrary set $\mathcal{A} \subseteq \mathcal{X}$ and computes the ratio of the probability of $\boldsymbol{x}$ over the sum of the probabilities of each configuration $\boldsymbol{x}'$ contained in the set $\mathcal{A}$.

We can now define the generalized estimation criterion function $f_\theta(\boldsymbol{x}_{1:N})$ as seen in Equation 3.10. We use the shorthand notation $\boldsymbol{x}_{1:N}$ to indicate a data set containing $N$ data cases $\boldsymbol{x}_1, ..., \boldsymbol{x}_N$. The estimation criterion is restricted to decompose additively across the $N$ data cases.

$$f_\theta(\boldsymbol{x}_{1:N}) = \frac{1}{N} \sum_{n=1}^{N} m_\theta(\boldsymbol{x}_n) \qquad (3.10)$$

$$m_\theta(\boldsymbol{x}) = \frac{1}{C} \sum_{c=1}^{C} g_c(R_\theta(\boldsymbol{x}, \mathcal{N}_c(\boldsymbol{x}))) \qquad (3.11)$$

For each data case, we apply the function $m_\theta(\boldsymbol{x})$ defined in Equation 3.11. $m_\theta(\boldsymbol{x})$ consists of a sum over $C$ components. For each component $c$, we compute the probability ratio $R_\theta(\boldsymbol{x}, \mathcal{N}_c(\boldsymbol{x}))$ and then pass it through a transfer function $g_c()$. The probability ratio is computed with respect to a set $\mathcal{N}_c(\boldsymbol{x}) \subseteq \mathcal{X}$ that we refer to as the neighborhood of $\boldsymbol{x}$ in component $c$. The neighborhood function can depend both on the component $c$ and the data configuration $\boldsymbol{x}$. The transfer function $g_c()$ can depend on the component $c$ only. When the number of components and the size of the neighborhoods are both relatively small, the resulting criterion function can be computed exactly and optimized without requiring any other approximation.

### 3.2 M-Estimators

Following van der Vaart [2000, Ch. 5], an M-estimator is defined through a criterion function that is repre-

$$\frac{\partial m_\theta(\boldsymbol{x})}{\partial \theta_i} = \frac{1}{C} \sum_{c=1}^{C} g'_c(R_\theta(\boldsymbol{x}, \mathcal{N}_c(\boldsymbol{x}))) R_\theta(\boldsymbol{x}, \mathcal{N}_c(\boldsymbol{x})) \left( -\frac{\partial E_\theta(\boldsymbol{x})}{\partial \theta_i} + \mathbb{E}_{R_\theta(\boldsymbol{x}, \mathcal{N}_c(\boldsymbol{x}))} \left[ \frac{\partial E_\theta(\boldsymbol{x})}{\partial \theta_i} \right] \right) \quad (3.6)$$

$$\frac{\partial^2 m_\theta(\boldsymbol{x})}{\partial \theta_i \partial \theta_j} = \frac{1}{C} \sum_{c=1}^{C} \left( g''_c(R_\theta(\boldsymbol{x}, \mathcal{N}_c(\boldsymbol{x}))) R_\theta^2(\boldsymbol{x}, \mathcal{N}_c(\boldsymbol{x})) + g'_c(R_\theta(\boldsymbol{x}, \mathcal{N}_c(\boldsymbol{x}))) R_\theta(\boldsymbol{x}, \mathcal{N}_c(\boldsymbol{x})) \right)$$

$$\cdot \left( -\frac{\partial E_\theta(\boldsymbol{x})}{\partial \theta_j} + \mathbb{E}_{R_\theta(\boldsymbol{x}, \mathcal{N}_c(\boldsymbol{x}))} \left[ \frac{\partial E_\theta(\boldsymbol{x})}{\partial \theta_j} \right] \right) \left( -\frac{\partial E_\theta(\boldsymbol{x})}{\partial \theta_i} + \mathbb{E}_{R_\theta(\boldsymbol{x}, \mathcal{N}_c(\boldsymbol{x}))} \left[ \frac{\partial E_\theta(\boldsymbol{x})}{\partial \theta_i} \right] \right)$$

$$+ \frac{1}{C} \sum_{c=1}^{C} g'_c(R_\theta(\boldsymbol{x}, \mathcal{N}_c(\boldsymbol{x}))) R_\theta(\boldsymbol{x}, \mathcal{N}_c(\boldsymbol{x})) \left( -\frac{\partial^2 E_\theta(\boldsymbol{x})}{\partial \theta_i \partial \theta_j} + \mathbb{E}_{R_\theta(\boldsymbol{x}, \mathcal{N}_c(\boldsymbol{x}))} \left[ \frac{\partial^2 E_\theta(\boldsymbol{x})}{\partial \theta_i \partial \theta_j} \right] \right)$$

$$- \frac{1}{C} \sum_{c=1}^{C} g'_c(R_\theta(\boldsymbol{x}, \mathcal{N}_c(\boldsymbol{x}))) R_\theta(\boldsymbol{x}, \mathcal{N}_c(\boldsymbol{x})) \mathbb{C}_{R_\theta(\boldsymbol{x}, \mathcal{N}_c(\boldsymbol{x}))} \left( \frac{\partial E_\theta(\boldsymbol{x})}{\partial \theta_j}, \frac{\partial E_\theta(\boldsymbol{x})}{\partial \theta_i} \right) \quad (3.7)$$

---

sentable as the empirical average of a known estimating function $m_\theta(\boldsymbol{x}_n)$ over $N$ samples $\boldsymbol{x}_n$ drawn from an underlying true probability distribution $P_*(\boldsymbol{x})$ as seen in Equation 3.12. Note that we can equivalently define the estimator as an expectation under the empirical distribution $P_N(\boldsymbol{x}) = \frac{1}{N} \sum_{n=1}^{N} \delta(\boldsymbol{x}, \boldsymbol{x}_n)$ where $\delta(a,b)$ is the Kronecker delta function. Our generalized estimator parameterizes the estimating function $m_\theta(\boldsymbol{x}_n)$, creating a useful restriction on the space of all M-estimators.

$$f_\theta(\boldsymbol{x}_{1:N}) = \frac{1}{N} \sum_{n=1}^{N} m_\theta(\boldsymbol{x}_n) = \sum_{\boldsymbol{x} \in \mathcal{X}} P_N(\boldsymbol{x}) m_\theta(\boldsymbol{x}) \quad (3.12)$$

The two key theoretical properties of M-estimators are asymptotic consistency and asymptotic efficiency. The advantage of working within the M-estimator framework is that generic conditions necessary for these properties to hold are well established.

### 3.3 Consistency

Following van der Vaart [2000, Ch. 5], we define $f_\theta^*$ to be the limit as $N$ goes to infinity of the criterion function $f_\theta(\boldsymbol{x}_{1:N})$, as given in Equation 3.13. We define $\theta^\infty$ to be an optimizer of $f_\theta^*$. Note that it is not necessarily assumed that the true distribution is in the model class so that $P_*$ is not in general equal to $P_{\theta^\infty}$.

$$f_\theta^* = \sum_{\boldsymbol{x} \in \mathcal{X}} P_*(\boldsymbol{x}) m_\theta(\boldsymbol{x}) \quad (3.13)$$

We denote an M-estimator computed by maximizing a criterion function $f_\theta(\boldsymbol{x}_{1:N})$ based on a sample of $N$ data cases drawn from $P_*$ by $\hat{\theta}^N$. The estimator $\hat{\theta}^N$ is said to be consistent if the limit as $N$ goes to infinity of $\mathcal{D}(\hat{\theta}^N - \theta^\infty)$ goes to zero under a suitable distance metric $\mathcal{D}$. All of the models we consider have parameters defined over Euclidean spaces, so we will simply take $\mathcal{D}$ to be the standard Euclidean distance. The requirements for an M-estimator to be consistent are given in van der Vaart [2000, Theorem 5.7]. Establishing the consistency of estimators for log-linear models can be relatively simple because the criterion functions are often convex. It is generally not possible to establish consistency for non-identifiable models like RBMs without placing further restrictions on the parameters.

### 3.4 Asymptotic Normality

An estimator is said to be asymptotically normal if it is consistent and the sequence $\sqrt{N}(\hat{\theta}^N - \theta^\infty)$ converges in distribution to a multivariate normal $\mathcal{N}(0, \Sigma)$. The conditions on the estimating function $m_\theta(\boldsymbol{x})$ required to ensure asymptotic normality are given in van der Vaart, Theorem 5.21. Our main interest is the asymptotic covariance matrix $\Sigma$, which can be defined in terms of $m_\theta(\boldsymbol{x})$ as shown in Equation 3.14.

$$\Sigma = (H_{\theta^\infty})^{-1} J_{\theta^\infty} (H_{\theta^\infty})^{-1} \quad (3.14)$$

$$(H_{\theta^\infty})_{ij} = E_{P_*(\boldsymbol{x})} \left[ \frac{\partial^2 m_\theta(\boldsymbol{x})}{\partial \theta_i \partial \theta_j} \bigg|_{\theta^\infty} \right]$$

$$(J_{\theta^\infty})_{ij} = E_{P_*(\boldsymbol{x})} \left[ \frac{\partial m_\theta(\boldsymbol{x})}{\partial \theta_i} \frac{\partial m_\theta(\boldsymbol{x})}{\partial \theta_j} \bigg|_{\theta^\infty} \right]$$

To define the asymptotic covariance matrix for our generalized estimator, we need only supply the first and second derivatives for our parametrization of the estimating function $m_\theta(\boldsymbol{x})$. The results are given in Equations 3.6 and 3.7. We use the notation $\mathbb{E}_{R_\theta(\boldsymbol{x}, \mathcal{N}_c(\boldsymbol{x}))}[a(\boldsymbol{x})]$ and $\mathbb{C}_{R_\theta(\boldsymbol{x}, \mathcal{N}_c(\boldsymbol{x}))}(a(\boldsymbol{x}), b(\boldsymbol{x}))$ to denote expectations and covariances under $R_\theta(\boldsymbol{x}, \mathcal{N}_c(\boldsymbol{x}))$. We define the expectation below. The definition of the covariance follows from the expectation in the usual way. These definitions are sensible since $R_\theta(\boldsymbol{x}, \mathcal{N}_c(\boldsymbol{x}))$ acts as a properly normalized dis-tribution over the set $\mathcal{N}_c(\boldsymbol{x})$.

$$\mathbb{E}_{R_\theta(\boldsymbol{x},\mathcal{N}_c(\boldsymbol{x}))}[a(\boldsymbol{x})] = \sum_{\boldsymbol{x}' \in \mathcal{N}_c(\boldsymbol{x})} R_\theta(\boldsymbol{x}', \mathcal{N}_c(\boldsymbol{x})) a(\boldsymbol{x}')$$

Despite appearing complex, Equations 3.6 and 3.7 can be derived with relative ease starting from Equation 3.11. More importantly, the equations depend only on the definition of the ratio function $R_\theta(\boldsymbol{x}, \mathcal{N}_c(\boldsymbol{x}))$, the first and second derivatives of the transfer function $g_c(R)$ and the energy function of the model $E_\theta(\boldsymbol{x})$. All of these quantities can be easily derived for specific models and estimators. In the next section, we show how several classical and recent estimators can be viewed as special cases of the generalized estimator.

## 4 Applications

Many classical and recently proposed estimators for dEBMs are subsumed by our generalized estimator through appropriate choices of the number of components $C$, the transfer function $g_c(R)$ and the neighborhood function $\mathcal{N}_c(\boldsymbol{x})$. To simplify the exposition, we focus on estimators for binary data and generative dEBMs only.

### 4.1 Maximum Likelihood

The maximum likelihood principle states that we should select the parameters $\theta$ that assign the highest probability to the observed data. The maximum likelihood criterion function is given in Equation 4.15.

$$f_\theta^{ML}(\boldsymbol{x}_{1:N}) = \frac{1}{N} \sum_{n=1}^{N} \log P_\theta(\boldsymbol{x}_n) \quad (4.15)$$

The maximum likelihood estimator can be put into our generalized framework using the settings $C = 1$, $g_1(R) = \log(R)$, and $\mathcal{N}_1(\boldsymbol{x}) = \mathcal{X}$ for all $\boldsymbol{x}$. This choice of neighborhood function exactly recreates the full partition function since $R_\theta(\boldsymbol{x}, \mathcal{N}_1(\boldsymbol{x})) = P_\theta(\boldsymbol{x})$ for all $\boldsymbol{x}$. The log transfer function has the interesting property that $\log'(R)R = 1$ and $\log''(R)R^2 = -1$. This cancels the entire first line of Equation 3.7. If we further assume that the model is well specified ($P_* = P_{\theta^\infty}$), we quickly arrive at the classical result that $H_{\theta^\infty}^{ML} = -J_{\theta^\infty}^{ML}$ and the asymptotic covariance matrix for maximum likelihood is equal to the inverse of the Fisher information matrix $\mathcal{I}$.[4]

$$\Sigma^{ML} = (H_{\theta^\infty}^{ML})^{-1} J_{\theta^\infty}^{ML} (H_{\theta^\infty}^{ML})^{-1} = (J_{\theta^\infty}^{ML})^{-1} = \mathcal{I}^{-1}$$

$$(J_{\theta^\infty}^{ML})_{ij} = \mathbb{C}_{P_*(\boldsymbol{x})} \left( \frac{\partial E_\theta(\boldsymbol{x})}{\partial \theta_j}, \frac{\partial E_\theta(\boldsymbol{x})}{\partial \theta_i} \right)$$

---
[4]Note that all expressions must be evaluated at $\theta = \theta^\infty$ after taking derivatives, as in Equation 3.14, but we suppress the explicit notation here and in subsequent expressions due to space limitations.

### 4.2 Pseudolikelihood

The classical pseudolikelihood criterion consists of a sum of all one-dimensional model log-conditional distributions $\log P_\theta(x_d|\boldsymbol{x}_{-d})$ as shown in Equation 4.16. The notation $\boldsymbol{x}_{-d}$ indicates the instantiations of all of the random variables except for $\boldsymbol{x}_d$.

$$f_\theta^{PL}(\boldsymbol{x}_{1:N}) = \frac{1}{ND} \sum_{n=1}^{N} \sum_{d=1}^{D} \log P_\theta(x_{dn}|\boldsymbol{x}_{-dn}) \quad (4.16)$$

Pseudolikelihood can be put into our generalized framework using the settings $C = D$, $\mathcal{N}_d(\boldsymbol{x}) = \{\boldsymbol{x}, \bar{\boldsymbol{x}}^d\}$, and $g(R) = \log(R)$. The notation $\bar{\boldsymbol{x}}^d$ indicates the instantiation formed by starting with $\boldsymbol{x}$ and flipping the value of $x_d$. This choice of neighborhood function leads to $R_\theta(\boldsymbol{x}, \mathcal{N}_d(\boldsymbol{x})) = P_\theta(x_d|\boldsymbol{x}_{-d})$, which recovers the conditional distributions used to define pseudolikelihood.

As in the maximum likelihood case, use of the log transfer function cancels the entire first line of Equation 3.7. If we again assume that the model is well specified ($P_* = P_{\theta^\infty}$), we can easily obtain a simple form for $(H_{\theta^\infty}^{PL})_{ij}$:

$$-\mathbb{E}_{P_*(\boldsymbol{x})} \left[ \frac{1}{D} \sum_{d=1}^{D} \mathbb{C}_{P_*(x_d|\boldsymbol{x}_{-d})} \left( \frac{\partial E_\theta(\boldsymbol{x})}{\partial \theta_j}, \frac{\partial E_\theta(\boldsymbol{x})}{\partial \theta_i} \right) \right] \quad (4.17)$$

The expression for $J_{\theta^\infty}^{PL}$ is somewhat more complicated. We state $(J_{\theta^\infty}^{PL})_{ij}$ below in an unexpanded form that is useful for subsequent comparisons.

$$\mathbb{E}_{P_*(\boldsymbol{x})} \left[ \frac{1}{D} \sum_{d=1}^{D} \left( \mathbb{E}_{P_*(x_d|\boldsymbol{x}_{-d})} \left[ \frac{\partial E_\theta(\boldsymbol{x})}{\partial \theta_i} \right] - \frac{\partial E_\theta(\boldsymbol{x})}{\partial \theta_i} \right) \right. $$
$$\left. \cdot \frac{1}{D} \sum_{d=1}^{D} \left( \mathbb{E}_{P_*(x_d|\boldsymbol{x}_{-d})} \left[ \frac{\partial E_\theta(\boldsymbol{x})}{\partial \theta_j} \right] - \frac{\partial E_\theta(\boldsymbol{x})}{\partial \theta_j} \right) \right] \quad (4.18)$$

### 4.3 Ratio Matching

The ratio matching criterion function can be interpreted as a sum of $\ell_2$ distances between pairs of one-dimensional conditional distributions under the model and the empirical distribution. The ratio matching criterion function is given in Equation 4.19. Note that the original presentation of ratio matching defines the estimator in terms of a function $h(a) = 1/(1+a)$ applied to ratios of the form $P(\boldsymbol{x})/P(\bar{\boldsymbol{x}}^d)$ [Hyvärinen, 2007], but this unnecessarily obscures the close relationship between ratio matching and pseudolikelihood.

$$f_\theta^{RM}(\boldsymbol{x}_{1:N}) = -\frac{1}{ND}\sum_{n=1}^{N}\sum_{d=1}^{D}\sum_{\xi\in\{0,1\}}\Big(P_N(X_d=\xi|\boldsymbol{x}_{-dn})$$
$$-P_\theta(X_d=\xi|\boldsymbol{x}_{-dn})\Big)^2 \quad (4.19)$$

The ratio matching criterion function can be reduced to a much simpler form as shown by Hyvärinen [2007]. We give this simplified form in Equation 4.20. Note that the extra summation over the states of the $d^{th}$ variable in Equation 4.19 is critical to obtaining this reduced form of the estimator shown below.

$$f_\theta^{RM}(\boldsymbol{x}_{1:N}) = -\frac{1}{ND}\sum_{n=1}^{N}\sum_{d=1}^{D}(1-P_\theta(x_{dn}|\boldsymbol{x}_{-dn}))^2 \quad (4.20)$$

Ratio matching can be obtained in our generalized framework by the settings $C=D$, $\mathcal{N}_d(\boldsymbol{x})=\{\boldsymbol{x},\bar{\boldsymbol{x}}^d\}$, and $g_d(R)=-(1-R)^2$. We immediately see that the choice of the number of components and the neighborhood structure is identical to pseudolikelihood. The two estimators differ only in terms of the transfer functions. We plot the $\log(R)$ transfer function as well as the $-(1-R)^2$ transfer function in Figure 1. We note that the two functions are very similar, differing mainly in the fact that $-(1-R)^2$ is equal to $-1$ at $R=0$ while $\log(R)$ diverges to $-\infty$ as $R$ goes to zero.

The fact that ratio matching uses a different transfer function means that the expression for the asymptotic covariance matrix is yet more complex. However, it does simplify under the assumption of no model mismatch. We obtain the following expression for $(H_{\theta\infty}^{RM})_{ij}$:

$$-\frac{2}{D}\sum_{d=1}^{D}\mathbb{E}_{P_*(\boldsymbol{x})}\Big[P_*^2(x_d|\boldsymbol{x}_{-d})$$
$$\cdot\Big(\mathbb{E}_{P_*(x_d|\boldsymbol{x}_{-d})}\Big[\frac{\partial E_\theta(\boldsymbol{x})}{\partial\theta_i}\Big]-\frac{\partial E_\theta(\boldsymbol{x})}{\partial\theta_i}\Big)$$
$$\cdot\Big(\mathbb{E}_{P_*(x_d|\boldsymbol{x}_{-d})}\Big[\frac{\partial E_\theta(\boldsymbol{x})}{\partial\theta_j}\Big]-\frac{\partial E_\theta(\boldsymbol{x})}{\partial\theta_j}\Big)\Big]$$

We state $(J_{\theta\infty}^{RM})_{ij}$ below. We introduce a new piece of notation $\mathbb{V}_{P_*}^d=(P_*(x_d|\boldsymbol{x}_{-d}))(1-P_*(x_d|\boldsymbol{x}_{-d}))$ to indicate the Bernoulli variance of the $d^{th}$ conditional distribution.

$$\mathbb{E}_{P_*(\boldsymbol{x})}\Big[\frac{2}{D}\sum_{d=1}^{D}\mathbb{V}_{P_*}^d\Big(\mathbb{E}_{P_*(x_d|\boldsymbol{x}_{-d})}\Big[\frac{\partial E_\theta(\boldsymbol{x})}{\partial\theta_i}\Big]-\frac{\partial E_\theta(\boldsymbol{x})}{\partial\theta_i}\Big)$$
$$\cdot\frac{2}{D}\sum_{d=1}^{D}\mathbb{V}_{P_*}^d\Big(\mathbb{E}_{P_*(x_d|\boldsymbol{x}_{-d})}\Big[\frac{\partial E_\theta(\boldsymbol{x})}{\partial\theta_j}\Big]-\frac{\partial E_\theta(\boldsymbol{x})}{\partial\theta_j}\Big)\Big]$$

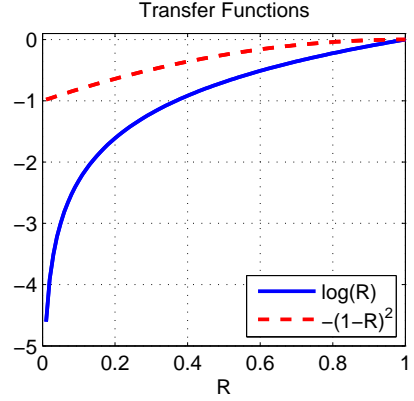

Figure 1: This figure shows the $\log(R)$ transfer function compared to the $-(1-R)^2$ transfer function used by ratio matching.

### 4.4 Generalized Score Matching

The inductive principle underlying generalized score matching also involves an $\ell_2$ distance and pairs of one-dimensional conditional distributions. In this case, however, the $\ell_2$ distance is applied to the difference of inverses conditional probabilities, as seen in Equation 4.21 [Lyu, 2009].

$$f_\theta^{GSM}(\boldsymbol{x}_{1:N}) = -\sum_{n=1}^{N}\sum_{d=1}^{D}\Big(\frac{1}{P_\theta(x_{dn}|\boldsymbol{x}_{-dn})}$$
$$-\frac{1}{P_N(x_{dn}|\boldsymbol{x}_{-dn})}\Big)^2 \quad (4.21)$$

The generalized score matching criterion can be reduced to a form that only depends on ratios of probabilities as shown in Equation 4.22.[5]

$$f_\theta^{GSM}(\boldsymbol{x}_{1:N}) = -\frac{1}{ND}\sum_{n=1}^{N}\sum_{d=1}^{D}\Big(\Big(\frac{P_\theta(x_{dn}|\boldsymbol{x}_{-dn})}{1-P_\theta(x_{dn}|\boldsymbol{x}_{-dn})}\Big)^{-2}$$
$$-2\Big(\frac{P_\theta(x_{dn}|\boldsymbol{x}_{-dn})}{1-P_\theta(x_{dn}|\boldsymbol{x}_{-dn})}\Big)\Big) \quad (4.22)$$

Generalized score matching can be put in our generalized framework using the settings $C=D$, $\mathcal{N}_d(\boldsymbol{x})=\{\boldsymbol{x},\bar{\boldsymbol{x}}^d\}$, and $g_d(R)=(R/(1-R))^{-2}-2(R/(1-R))$. In the interest of space, we omit the form of the asymptotic covariance matrix for generalized score matching as we will not analyze it further in the current paper.

### 4.5 Contrastive Estimation and Non-Local Contrastive Objectives

Contrastive estimation [Smith and Eisner, 2005] and non-local contrastive objectives [Vickrey et al., 2010]

---

[5]Note that the original reduced form due to Lyu contained an error that was corrected in Marlin et al. [2009]

are both closely related to composite conditional likelihood estimators [Lindsay, 1988]. Both estimators use multiple components, $\log(R)$ for the transfer function and ratios of unnormalized probabilities defined by neighborhoods. The main contribution of the contrastive estimation method is that it makes use of relatively large, structured neighborhoods that can be summed over efficiently.

The main feature of the contrastive objectives method is the proposal of a method for iteratively increasing the size of the neighborhoods. The goal of the iterative procedure is to dynamically construct neighborhoods where the configurations are not one-neighbors as in psuedolikelihood. In the interest of space, we again omit the forms of the asymptotic covariance matrix for these methods as we will not analyze them further in the current paper.

## 5 Relative Efficiency Results

It is well known that maximum likelihood attains the minimum possible asymptotic covariance of any consistent estimator. Prior work has considered the relative efficiency of pseudolikelihood compared to maximum likelihood [Liang and Jordan, 2008]. Our main interest in this work is establishing results regarding the relative efficiency of pseudolikelihood and ratio matching. It is well known that pseudolikelihood is consistent for well specified log-linear models since the criterion function is convex. The ratio matching criterion function is not convex, but it is still consistent for identifiable models in the well-specified case as shown by Hyvärinen [2007].

Our main theorem shows that the asymptotic covariance matrix for ratio matching can be both upper and lower bounded by scalar multiples of the pseudolikelihood asymptotic covariance matrix in the well-specified case. The inequalities in the theorem are with respect to the standard positive definite ordering relation where $A \preccurlyeq B$ implies that $B - A$ is positive definite.

**Theorem 5.1.** *Let $\mathbb{V}_{\max}$ and $\mathbb{V}_{\min}$ be the maximum and minimum values of the Bernoulli variance $P_*(x_d|\boldsymbol{x}_{-d})(1 - P_*(x_d|\boldsymbol{x}_{-d}))$ over all $\boldsymbol{x} \in \mathcal{X}$ and $d \in \{1...D\}$. Let $q_{\max}$ and $q_{\min}$ be the maximum and minimum values of $P_*(x_d|\boldsymbol{x}_{-d})$ over all $\boldsymbol{x} \in \mathcal{X}$ and $d \in \{1...D\}$. The asymptotic covariance matrix for ratio matching satisfies:*

$$\frac{\mathbb{V}_{min}^2}{q_{\max}^4}\Sigma^{PL} \preccurlyeq \Sigma^{RM} \preccurlyeq \frac{\mathbb{V}_{max}^2}{q_{\min}^4}\Sigma^{PL}$$

*Proof:* We begin with $J_{\theta^\infty}^{RM}$. We can easily see that the only difference between $J_{\theta^\infty}^{RM}$ and $J_{\theta^\infty}^{PL}$ is that each term in $J_{\theta^\infty}^{RM}$ is weighted by $2\mathbb{V}_{P_*}^d = 2P_*(x_d|\boldsymbol{x}_{-d})(1 - P_*(x_d|\boldsymbol{x}_{-d}))$. Since $\mathbb{V}_{\max}$ and $\mathbb{V}_{\min}$ upper and lower bound $\mathbb{V}_{P_*}^d$ by definition, we have that

$$4\mathbb{V}_{\min}^2 J_{\theta^\infty}^{PL} \preccurlyeq J_{\theta^\infty}^{RM} \preccurlyeq 4\mathbb{V}_{\max}^2 J_{\theta^\infty}^{PL}.$$

Under the assumption that $P_*(x_d|\boldsymbol{x}_{-d})$ is equal to a constant $q$, $H_{\theta^\infty}^{RM}$ simplifies $2q^2 H_{\theta^\infty}^{PL}$. Since $q_{\max}$ and $q_{\min}$ upper and lower bound $P_*(x_d|\boldsymbol{x}_{-d})$ by definition and $-H_{\theta^\infty}^{RM}$ and $-H_{\theta^\infty}^{PL}$ are positive definite, we have that $-2q_{\min}^2 H_{\theta^\infty}^{PL} \preccurlyeq -H_{\theta^\infty}^{RM} \preccurlyeq -2q_{\max}^2 H_{\theta^\infty}^{PL}$. By inverting all of the matrices we obtain:

$$-\frac{1}{2q_{\max}^2}(H_{\theta^\infty}^{PL})^{-1} \preccurlyeq -(H_{\theta^\infty}^{RM})^{-1} \preccurlyeq -\frac{1}{2q_{\min}^2}(H_{\theta^\infty}^{PL})^{-1}.$$

We can combine the two sets of bounds to prove the theorem as shown below.

$$\begin{aligned}\frac{\mathbb{V}_{\min}^2}{q_{\max}^4}\Sigma^{PL} &= \frac{\mathbb{V}_{\min}^2}{q_{\max}^4}(H_{\theta^\infty}^{PL})^{-1} J_{\theta^\infty}^{PL}(H_{\theta^\infty}^{PL})^{-1} \\ &\preccurlyeq \Sigma^{RM} = (H_{\theta^\infty}^{RM})^{-1} J_{\theta^\infty}^{RM}(H_{\theta^\infty}^{RM})^{-1} \\ &\preccurlyeq \frac{\mathbb{V}_{\max}^2}{q_{\min}^4}\Sigma^{PL} = \frac{\mathbb{V}_{\max}^2}{q_{\min}^4}(H_{\theta^\infty}^{PL})^{-1} J_{\theta^\infty}^{PL}(H_{\theta^\infty}^{PL})^{-1} \quad \square\end{aligned}$$

Since the coefficients in the bound are constructed by minimizing and maximizing over all conditional probabilities and variances, they are not independent. In particular, we must have that $q_{\max} = 1 - q_{\min}$. The minimal conditional variance $\mathbb{V}_{\min}$ is obtained using the most extreme conditional probabilities. This means that $\mathbb{V}_{\min} = q_{\min}(1 - q_{\min}) = (1 - q_{\max})q_{\max} = q_{\min}q_{\max}$. The maximal variance $\mathbb{V}_{\max}$ is independent of the other quantities, but can be expressed in terms of the conditional probability that is closest to 0.5 without exceeding it. We label this probability $q_{\text{mid}}$ and express the maximal variance as $\mathbb{V}_{\max} = q_{\text{mid}}(1 - q_{\text{mid}})$. This allows us to re-express the bound as follows:

$$\frac{q_{\min}^2}{q_{\max}^2}\Sigma^{PL} \preccurlyeq \Sigma^{RM} \preccurlyeq \frac{q_{\text{mid}}^2(1 - q_{\text{mid}})^2}{q_{\min}^4}\Sigma^{PL}$$

Since $q_{\min} \leq q_{\max}$, we have that $q_{\min}^2/q_{\max}^2 \leq 1$. Since $q_{\min} \leq q_{\text{mid}} \leq (1 - q_{\text{mid}})$, we have that $q_{\text{mid}}^2(1 - q_{\text{mid}})^2/q_{\min}^4 \geq 1$. As a result, we have that the same bound applies to the pseudolikelihood covariance matrix as well:

$$\frac{q_{\min}^2}{q_{\max}^2}\Sigma^{PL} \preccurlyeq \Sigma^{PL} \preccurlyeq \frac{q_{\text{mid}}^2(1 - q_{\text{mid}})^2}{q_{\min}^4}\Sigma^{PL}$$

We now consider the problem of quantifying the difference between $\Sigma^{RM}$ and $\Sigma^{PL}$. One standard way of comparing the size of covariance matrices is the ratio of their determinants. The determinant of a positive definite matrix is the volume of the corresponding ellipsoid. The determinant has a very intuitive interpretation in this setting: it can be thought of as the volume of parameter space in which the estimated parameters are likely to fall for large $N$. In the present case, a more convenient measure is the log of the determinant ratio, which is simply the difference of log determinants. Letting $l = \frac{\mathbb{V}_{min}^2}{q_{max}^4} = \frac{q_{min}^2}{q_{max}^2}$ and $h = \frac{\mathbb{V}_{max}^2}{q_{min}^4} = \frac{q_{mid}^2(1-q_{mid})^2}{q_{min}^4}$, we obtain the result:

$$\log\left(\frac{|l\Sigma^{PL}|}{|\Sigma^{PL}|}\right) \leq \log\left(\frac{|\Sigma^{RM}|}{|\Sigma^{PL}|}\right) \leq \log\left(\frac{|h\Sigma^{PL}|}{|\Sigma^{PL}|}\right)$$
$$D_\theta \log(l) \leq \log|\Sigma^{RM}| - \log|\Sigma^{PL}| \leq D_\theta \log(h)$$

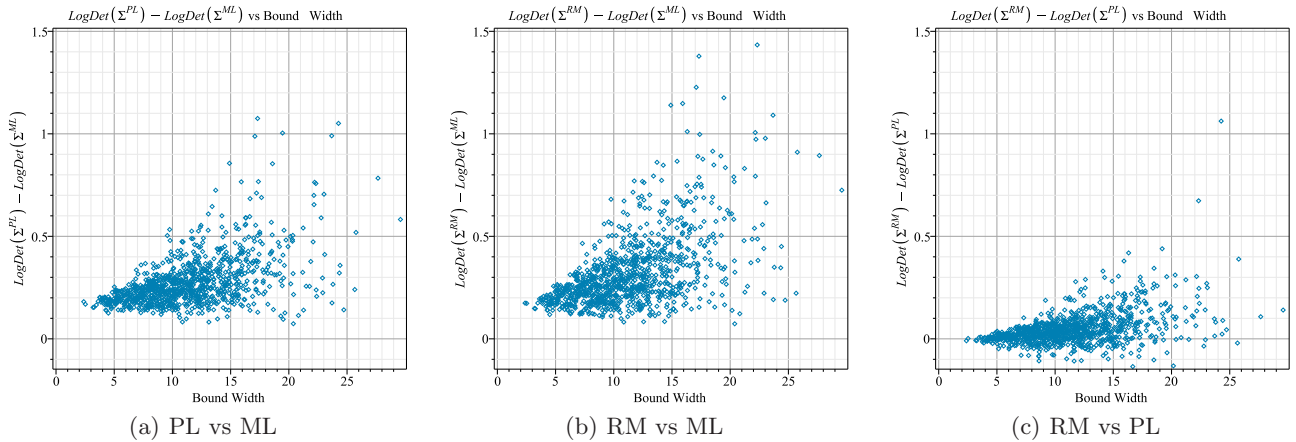

Figure 2: Efficiency difference for PL&ML (a), RM&ML (b) and RM&PL (c) versus the bound width.

We see that the width of the bound on the difference in asymptotic efficiency between pseudolikelihood and ratio matching in the log determinant sense depends directly on the width of the interval $\log(h) - \log(l)$. We note that in certain special cases the bound can be exact. For example, if the true distribution is uniform, all the conditional probabilities will be equal to $1/2^{|\mathcal{X}|}$. In this case, the covariance matrices for pseudolikelihood and ratio matching will be identical and the width of the bound in the above sense will also be exactly zero. On the other hand, the maximum width of the bound is unbounded as $q_{min}$ goes to zero. For an energy-based model, this would require that a subset of the parameter values approach positive or negative infinity.

### 5.1 Simulations

We perform a simulation study in low dimensions where the exact computation of the partition function as well as the asymptotic covariance matrices is feasible. The simulation uses a third-order binary Boltzmann machine in three dimensions with the following energy function: $E_\theta(\boldsymbol{x}) = -(W_1(1-x_1)(1-x_2)(1-x_3) + W_2 x_1(1-x_2)(1-x_3) + W_3(1-x_1)x_2(1-x_3) + W_4(1-x_1)(1-x_2)x_3 + W_5 x_1 x_2(1-x_3) + W_6 x_1(1-x_2)x_3)$. This energy function thus has six parameters while the full joint distribution has seven degrees of freedom.

We randomly sample 1000 sets of true model parameters from a zero mean spherical Gaussian distribution with standard deviation one. We then compute the determinant of the asymptotic covariance matrix for maximum likelihood, pseudolikelihood and ratio matching, as well as the bound coefficients $l = \frac{\mathbb{V}_{min}^2}{q_{max}^4}$ and $h = \frac{\mathbb{V}_{max}^2}{q_{min}^4}$. We plot the log determinant differences for each set of sampled parameters relative to the bound width $\log(h) - \log(l)$ for each pair of estimators.

The results are given in Figure 2 for PL and ML (a), RM and ML (b) and RM and PL (c). In each plot, the second estimator in the pair is *more efficient* for a given sample in the log-determinant sense if the plotted point lies above zero (this indicates that the second estimator in the pair has an asymptotic covariance matrix with *smaller* volume). As expected, we see that ML is always more efficient than both PL and RM in the log-determinant sense. As predicted by the theoretical development in the previous section, the maximum difference in efficiency between ratio matching and pseudolikelihood in the log-determinant sense scales with the bound width, although we see that the bound can be quite loose. The plots clearly show that neither estimator is always more efficient than the other in the log-determinant sense. We also note that neither is more efficient than the other in the stricter positive definite ordering. Indeed, $\Sigma^{RM} - \Sigma^{PL}$ is almost always an *indefinite* matrix with some positive and some negative eigenvalues. For example, the parameter setting $W = [2, 2, 2, -2, -2, -2]$ leads to $\Sigma^{RM} - \Sigma^{PL}$ having the eigenvalues: $[10.8, 0, 0, 0, 0, -6.5]$.

## 6 Conclusions and Future Work

We present a generalized estimator that unifies several classical and recently proposed estimators. This uni-fication is valuable for several reasons. First, it highlights the small number of dimensions along which the estimators differ: the choice of transfer function, the number of components and the neighborhood structure. Second, it allows us to derive generic asymptotic results by applying M-estimation theory to the generalized estimator. In this paper, we use these results to study the relative asymptotic efficiency of pseudolikelihood and ratio matching.

The unifying perspective offered by the generalized estimator also raises many interesting questions for future study. For example, can we obtain general results regarding the relative efficiency of estimators as a function of properties of the transfer function and neighborhood structure? Can we design transfer functions that are maximally efficient for a given neighborhood structure and model class? How does the choice of transfer function affect robustness of the estimators? Alternative estimators with a large number of components will again necessitate the use of stochastic approximation algorithms. The question of how the efficiency of such estimators compares to the efficiency of stochastic maximum likelihood estimators is also of great interest.

## Acknowledgments

This work was supported by the Pacific Institute for the Mathematical Science, Mitacs, the CIFAR Neural Computation and Adaptive Perception program and the Natural Sciences and Engineering Research Council of Canada.